\documentclass[11pt]{article}
\usepackage[final]{emnlp2021}
\usepackage{times}
\usepackage[utf8]{inputenc}
\usepackage[T1]{fontenc}
\usepackage{url} 
\usepackage{latexsym} 
\usepackage{subcaption}
\usepackage{color}

\title{
Language models in word sense disambiguation for Polish}

\author{Agnieszka Mykowiecka,   Agnieszka A. Mykowiecka,  
\and Piotr Rychlik\\
Institute of Computer Science Polish Academy of Sciences, \\Jana Kazimierza 5, Warsaw, Poland} 

\date{}

\begin{document}
\maketitle
\begin{abstract}
In the paper, we test two different approaches to the {unsupervised}  word sense disambiguation task for Polish. 
In both methods, we use neural language models to predict words similar to those being disambiguated and, on the basis of these words, we predict the partition of word senses in different ways. In the first method, we cluster selected similar words, while in the second, we cluster vectors representing their subsets. The evaluation was carried out on texts annotated with plWordNet senses and provided a relatively good result (F1=0.68 for all ambiguous words). The results are significantly better than those obtained for the neural model-based unsupervised method proposed in \cite{waw:myk:17:Sense} and are at the level of the supervised method presented there. 
The proposed method   may be a way of solving  word sense disambiguation problem for languages  which lack sense annotated data.
\end{abstract}

\section{Introduction}
The problems of word sense disambiguation (WSD) and discrimination for Polish have not been a popular area of research. The main reason is probably the lack of annotated data that can be used for evaluation purposes and as training data in supervised methods. Another reason may be that disambiguating and discriminating between the senses of words is widely regarded as a daunting challenge, especially as the results achieved in this field are not satisfactory, even for the English language.

Mihalcea et al. in \shortcite{Mihalcea:2004} claimed that their results from the WSD task, which achieved an F1 score of 0.65 (and were the best at the time),  were at the human level according to the inter-annotator agreement.
However, in recent years, more efficient results for English were published. For example,  Yuan et al.\
\shortcite{yuan-etal-2016-semi}  reported the LSTM based results with an F1 score between 0.69 and 0.82, depending on the training set. Similarly, Reganato et al.\ 
\shortcite{raganato-etal-2017-neural} achieved an F1 of about 0.8 using LSTM encoder-decoder architecture trained on the SemCor 3.0 English data set annotated with WordNet senses. Supervised methods, therefore, turned out to be quite effective in this task, at least for English. However, the use of such methods requires large annotated data sets because we need to have examples of the use of words in every sense that we would like to identify. To solve this problem, a semi-supervised method for WSD was proposed in \cite{yarowsky-1995-unsupervised}, where only a small number of examples were necessary. Strictly unsupervised approaches, where no annotated data is required for training, were tried next.  The most frequent and efficient solutions used semantic knowledge nets like WordNet. For example, Agirre et al.\ \shortcite{eneco:06} presented new parameters' tuning for PageRank based graph solutions and achieved an F1 equal to 0.64. These early approaches were summarized in \cite{agi:edm:2006}. In the more recent Babelfly solution \cite{babelfly:14}, a method based on using the densest subgraph of a semantic network used to select a high-coherence semantic interpretation of a text gave an F1 equal to 0.66, 0.87, and  0.69 when using WordNet, Wikipedia, and BabelNet English semantic resources respectively. The modified random walk over large lexical knowledge bases proposed in \cite{agirre:14} did not rely on sense frequency data, which is not available for many languages, and it outperformed the most frequent sense method for Spanish.

The work on unsupervised WSD in Polish also began with using graph-based methods which were presented in \cite{kedzia2015word} and \cite{janz:pias:2019}.  In the latter paper, the personalized PageRank method was tested on several different knowledge bases  and test sets with the precision reported at very different levels from 0.4 to 0.7.
After the emergence of new distributed language models, other methods not relying on knowledge bases were developed. Wawer and Mykowiecka \shortcite{waw:myk:17:Sense}  proposed a method based on comparing sentence probability counted using pre-trained word embeddings of unambiguous word synonyms from all synsets it belongs to. The method was tested on data from \cite{hajn:14b}.  
 In \cite{ourgwc:2020}, to estimate the likelihood of a sense appearing amidst the word sequence,
the token being disambiguated was substituted with words lexically related to
the given sense or words appearing in its
WordNet gloss. None of these efforts gave results better than the F1 score of approximately 0.6.

In this work, we propose two solutions for the word sense disambiguation task for Polish using different neural language models. To eliminate the need for collecting a large training set, we treat the problem as a word sense discrimination task, i.e. we cluster word occurrences in groups which presumably correspond to different senses. After that, we use a small annotated subset of these occurrences to assign groups to plWordNet senses. The evaluation is performed using the same scenario, i.e. we evaluate the results on the annotated subset of word occurrences. In the first method,  we cluster Word2Vec embeddings of words similar to words to be disambiguated. We then assign a given  occurrence of a word to the cluster based on its similarity to the context of the word.
In the second, based on \cite{amrami-goldberg-2018-word}, we use the BERT masked  model  to predict words which can replace a given one or can co-occur with it in a conjunctive phrase. The sets of possible replacements, represented as vectors, are clustered to find out their semantic similarity. While evaluating the results, we focused not on the overall results for all annotated word occurrences, but on the results obtained for selected words with more than two senses in our test set. 

\section{Data Description}

We performed our clustering experiments on the NKJP corpus (\textit{National Corpus of Polish}, \cite{nkjp11a}). Unfortunately, this data set is not annotated with word senses,  so to check the quality of our methods we had to rely on its small subset which was annotated within the project -- an extended version of the one used in \cite{ourgwc:2020} available from xxx\footnote{The exact version of the set and the program code will be available if the paper is accepted.}. This dataset consists of about 90,000 tokens among which there are about 4,300 different nouns, verbs and adjectives manually annotated by appropriately trained linguists
with plWordNet \cite{wordnet:19} version 3.2 sense identifiers. The inter-annotator agreement was calculated only on the randomly selected subset of data, and two annotators agreed on 83\% of the tokens, which is quite a good result for this task (significantly better than the already mentioned threshold given in \cite{Mihalcea:2004} for English).  As we were interested in identifying different word senses, we decided to look not only at the overall results, but also into the detailed data to find out how the algorithm performs on the words which are ambiguous not only potentially, but actually within our test data. The selected words divided into syntactic categories -- words for which there are at least four occurrences representing (in total) at least two senses, are shown in Table \ref{tab:words}. We can see that the maximal frequency is equal to 33,  the maximal different sense count is 9, while the most typical number of senses in this set is four (13 cases for 30 words). The proportion of the number of senses which actually occurred in the set to the number of possible senses identified in plWordNet varies from below 0.2 (3 out of 19 senses for \textit{block} and 4 from 22 senses for \textit{line)} to 1 for \textit{office} and \textit{operate}).

 \begin{table}[ht]
  \setlength{\tabcolsep}{5pt}
 \begin{small}
 \begin{tabular}{r|l|l|r|rrr|rrll}
 \hline
 & {\bf word} & {\bf English} & \#  & \multicolumn{2}{c}{\bf \# of senses} \\
 & & {\bf translation}&& {\bf plWordNet} & {\bf data}\\
  \hline
  & \multicolumn{2}{l|}{\sc nouns}&&&\\
1&badanie   &  examination &13  & 6 & 4\\
2&biuro   &  office &14  & 3 & 3\\
3&blok   &  block & 5  & 19 & 3\\
4&głos   &  voice/vote & 21  & 14 & 4\\
5&historia  &  history & 19  & 6 & 4\\
6&interes   &  business/deal & 14  &  5 & 4\\
7&język   &  language & 11  &  10 & 3\\
8&kierunek   &  direction & 13  &  5 & 3\\
9&klasa   &  class & 7  &   14 & 4\\
10&kolej   &  rail/turn &4  &  5 & 3\\
11&koło   &  circle/wheel & 12  &  12 & 4\\
12&komórka   &  cell/niche &7  &  9 & 3 \\
13&linia   &  line & 5  &  22 & 4 \\
14&zamek   &  castle/lock &6  & 7 & 2\\
\hline
  & \multicolumn{2}{l|}{\sc adjectives}&&&\\
15&biały   &  white & 16 & 17 & 6 \\
16&ciężki   &  heavy & 15  &   7 & 6\\
17 &łagodny  &  mild & 7 & 18& 5\\
18&niski  &  low & 11 & 11& 5\\
19&prawdziwy  & real & 11 & 8 & 4\\
20&szybki  &  fast & 18  &9& 4\\
21&zły & bad & 17 & 11 &  6\\
\hline
 & \multicolumn{2}{l|}{\sc verbs}&&&\\
22&działać  &  operate & 18 &5& 5\\
23&należeć  &   belong& 26& 6& 4\\
24&przyjąć  &  accept & 24 &13 & 9\\
25&uważać  & regard & 33 & 5 & 2\\
26&zakładać  &  assume& 5 &11 & 3\\
27&zamykać  &    close & 10 &11& 4\\
28&zobaczyć  & see & 8 &10 & 4\\
29&związać  &   bind & 7 &11 & 2 \\
30&żyć  & live & 13 &6 & 4\\
\hline
\end{tabular}
\caption{ List of words for which detailed evaluation was performed with the numbers of: word occurrences within the annotated data,  word senses identified in plWordNet and different senses observed within the annotated data. There are 390 occurrences of these words in total.}
\label{tab:words}
\end{small}
\end{table}

\section{Direct Use of Word Embeddings}

Our first solution is a naive method based on the assumption that Word2Vec  word embeddings somehow encode different meanings of words. We used Polish models from  \cite{myk:mar:ryc:17:CS}, precisely, the CBOW model\footnote{\tt nkjp+wiki-lemmas-all-300-cbow-ns-50} with vectors of size 300 trained on lemmatized NKJP and Polish Wikipedia using the negative sampling method and reduced to words occurring at least 50 times in the entire corpus. This model is available at \url{http://dsmodels.nlp.ipipan.waw.pl/}.

The exact meaning of a particular occurrence of an ambiguous word is defined by its context. In our first approach, the vector representation of a word occurrence is derived from the Word2Vec vectors of all its context words. This vector is then mapped to one of the clusters identified among the words similar to the word we want to disambiguate.

The method consists of two steps. In the first one, for each hand-tagged word in our database, we collected a set of the vectors that were most similar in our Word2Vec model according to the cosine distance. We assumed that each of these sets was likely to be sufficiently numerous and contain at least 100 items. Moreover, the elements of this set should contain words that are relatively strongly related (a similarity level of at least 0.4) to the corresponding database word. However, about 40\% of words in our database did not meet these conditions. In these cases, we have simply collected the 100 most similar words. The average size of the sets collected is 302 items. 

For each set obtained in such a way, we constructed a graph in which all vector pairs with the cosine similarity of at least 0.4 are connected. The nodes in these graphs were then clustered  using the Chinese whispers algorithm.
Clusters are assumed to represent the senses of the corresponding words.

In the next step, all considered words were assigned to one of the generated clusters based on their context vectors. This assignment was performed in two ways. For a given word, we calculated the average vector from the Word2Vec vectors of words belonging to the context of that word (up to 5 words on the left and right side). If the context was too short (less than 4), we did not consider such a word and did not assign it to any cluster.
We then compared this average context vector with the word vectors in the clusters that were generated in the previous step. The first proposed method consisted in selecting the cluster with the highest average similarity of its vectors to the average context vector. In the second one, we chose the cluster that contained a word whose vector was closest to the average context vector. 

The results of both these approaches for the set of selected 30 words are given in Table \ref{tab:1}.  We use standard precision and recall counted for every cluster member (Bcubed evaluation measure, \cite{bcubed:09}). We also tested the above methods on the entire data set containing about 4,350 different annotated words. For all ambiguous words in the test set counted together, we obtained a higher result than those shown in Table \ref{tab:1} for the selected words. An F1 score reached a level of 0.68. This is not surprising as most of the words in our data set are much less ambiguous than the words out of the selected 30. When we also considered words that are unambiguous, an F1 score increased to 0.8.

In spite of the fact that processing time increases linearly with the number of nodes for the Chinese whispers algorithm, clustering is the most computationally demanding part of the presented method.
However, it is possible to create a cluster database for all the vocabulary of the language model used and then apply that database to different datasets.

 \begin{table*}[ht!]
  \setlength{\tabcolsep}{5pt}
 \begin{small}
 \begin{tabular}{rl|rrr|rrr||rrr|rrrll}
 \hline
 && \multicolumn{12}{c}{\bf Method I: Direct use of word embeddings}\\
 && \multicolumn{6}{c||}{\bf average similarity}& \multicolumn{6}{c}{\bf maximal similarity}\\
  & &     P$_n$  & R$_n$  &  F1$_n$  &  P$_{nc}$  &  R$_{nc}$  &  F1$_{nc}$    &     P$_n$  & R$_n$  &  F1$_n$  &  P$_{nc}$  &  R$_{nc}$  &  F1$_{nc}$   \\
  \hline
1&badanie  &  0.28  &  1  &  {0.44}  &  0.28  &  1  & {0.44} &  0.28  &  1  &  {0.44}  &  0.28  &  1  &  \bf{0.44}\\
2&biuro  &  0.67  &  0.52  & \fbox{{\bf  0.59}}  &  0.80  &  0.42  &  0.55   &  0.59  &  0.45  &  0.51  &  0.73  &  0.37  &  0.49\\
3&blok  &  0.36  &  1  &  {0.53}  &  0.36  &  1  &  {0.53}   &  0.36  &  1  &  {0.53}  &  0.36  &  1  &  {0.53}\\
4&głos  &  0.54  &  0.81  &  0.65  &  0.79  &  0.59  &  \fbox{\underline{\bf 0.68}}   &  0.46  &  0.54  &  0.50  &  0.58  &  0.38  &  0.46\\
5&historia  &  0.53  &  0.89  &  \fbox{\bf 0.66}  &  0.54  &  0.80  &  0.65   &  0.41  &  1  &  0.58  &  0.41  &  1  &  0.58\\
6&interes  &  0.60  &  0.50  &  0.54  &  0.66  &  0.35  &  0.46  &  0.54  &  0.86  &  \fbox{\underline{\bf{0.66}}}  &  0.75  &  0.52  &  0.61\\
7&język  &  0.46  &  1  &  \fbox{{0.63}}  &  0.46  &  1  &  \fbox{{0.63}} &  0.46  &  1  &  \fbox{{0.63}}  &  0.46  &  1  &  \fbox{{0.63}}\\
8&kierunek  &  1  &  0.37  &  0.54  &  1  &  0.43  &  \underline{0.60}  &  0.90  &  0.47  &  0.62  &  0.89  &  0.57  &  \bf{\underline{0.69}}\\
9&klasa  &  0.67  &  0.67  &  0.67  &  0.76  &  0.67  &  \bf\underline{0.72} &  0.86  &  0.57  &  0.69  &  0.92  &  0.60  &  \bf\underline{0.72}\\
10&kolej  &  0.75  &  0.75  &  0.75  &  0.83  &  0.75  &  \bf\underline{0.79} &  0.75  &  0.75  &  0.75  &  0.83  &  0.75  &  \bf\underline{0.79}\\
11&koło  &  0.72  &  0.57  &  \bf{0.64}  &  0.71  &  0.59  &  \bf{0.64}  &  0.68  &  0.53  &  0.59  &  0.74  &  0.47  &  0.58\\
12&komórka  &  0.33  &  1  &  0.50  &  0.33  &  1  &  0.50 &  0.33  &  1  &  0.50  &  0.33  &  1  &  0.50\\
13&linia  &  0.28  &  1  &  0.44  &  0.28  &  1  &  0.44  &  0.28  &  1  &  0.44  &  0.28  &  1  &  0.44\\
14&zamek  &  0.52  &  1  &  0.68  &  0.52  &  1  &  0.68 &  0.52  &  1  &  0.68  &  0.52  &  1  &  0.68\\
\hline
& {\sc avg$_{noun}$} &0.55&	0.79&	0.50&	0.59&	0.76&	0.55&	0.53&	0.80&	0.58&	0.58&	0.76&	0.58\\
\hline
15&biały  &  0.88  &  1  &  0.94  &  0.88  &  1  &  0.94 &  0.88  &  1  &  0.94  &  0.88  &  1  &  \fbox{\bf{0.94}}\\
16&ciężki  &  0.27  &  0.76  &  0.39  &  0.29  &  0.62  &  \bf{0.40}&  0.25  &  0.68  &  0.37  &  0.29  &  0.60  &  0.39\\
17&łagodny  &  0.60  &  0.80  &  0.69  &  0.78  &  {0.78}  &   \bf\underline{0.78} &  0.28  &  1  &  0.44  &  0.28  &  1  &  0.44\\
18&niski  &  0.55  &  0.77  &  0.64  &  0.58  &  0.77  &  \fbox{\bf{0.66}} &  0.24  &  1  &  0.39  &  0.24  &  1  &  0.39\\
19&prawdziwy  &  0.35  &  0.80  &  0.49  &  0.41  &  0.69  &  0.51 &  0.26  &  1  &  0.41  &  0.26  &  1  &  0.41\\
20&szybki  &  0.48  &  0.41  &  0.45  &  0.49  &  0.40  &  0.44  &  0.54  &  0.72  &  0.62  &  0.70  &  0.64  & \fbox{\underline{\bf{0.67}}} \\
21&zły  &  0.34  &  0.90  &  0.49  &  0.65  &  0.56  & \fbox{\underline{\bf{0.60}}} &  0.33  &  0.73  &  0.46  &  0.35  &  0.64  &  0.45\\
\hline
& {\sc avg$_{adj}$} & 0.50&	0.78&	0.58&	0.58&	0.69&	\fbox{\bf{0.62}}&	0.40&	0.88&	0.52&	0.43&	0.84&	0.53
\\
\hline
22&działać  &  0.62  &  0.46  &  0.53  &  0.64  &  0.39  &  0.48  &  0.59  &  0.58  &  \underline{\bf{0.58}}  &  0.73  &  0.37  &  0.49\\
23&należeć  &  0.39  &  0.74  &  \underline{\bf{0.51}}  &  0.39  &  0.53  &  0.45 &  0.41  &  0.50  &  0.45  &  0.46  &  0.39  &  0.42\\
24&przyjąć  &  0.52  &  0.74  &  0.61  &  0.68  &  0.79  &  \fbox{\underline{\bf{0.73}}} &  0.36  &  1  &  0.53  &  0.36  &  1  &  0.53 \\
25&uważać  &  0.69  &  1  &  \fbox{0.81}  &  0.69  &  1  &  \fbox{0.81} &   0.69  &  1  &  \fbox{0.81}  &  0.69  &  1  &  \fbox{0.81}\\
26&zakładać  &  0.44  &  1  &  0.61  &  0.44  &  1  &  0.61 &  0.50  &  0.73  &  0.59  &  0.69  &  0.58  &  \bf{0.63}\\
27&zamykać  &  0.28  &  1  &  0.44  &  0.28  &  1  &  0.44 &  0.28  &  1  &  0.44  &  0.28  &  1  &  0.44\\
28&zobaczyć  &  0.42  &  1  &  0.59  &  0.42  &  1  &  0.59 &  0.42  &  1  &  0.59  &  0.42  &  1  &  0.59\\
29&związać  &  0.53  &  1  &  \fbox{0.69}  &  0.53  &  1  &  \fbox{0.69} &  0.53  &  1  &  \fbox{0.69}  &  0.53  &  1  &  \fbox{0.69}\\
30&żyć  &  0.64  &  0.50  &  0.56  &  0.67  &  0.44  &  0.53 &  0.73  &  0.63  &  \bf{0.68}  &  0.81  &  0.58  &  \bf{0.68}\\
\hline
& \sc{avg$_{verb}$} & 0.50&	0.83&	0.59&	0.53&	0.79&	0.59&	0.50&	0.83&	\bf{0.60}&	0.55&	0.77&	0.59\\
\hline
&{\sc avg}$_{30}$ &0.52  &  0.80  &  0.59  &  0.57  &  0.75  &  \bf{0.60} &0.49  &  0.82  &  0.57  &  0.53  &  0.78  &  0.57\\
\hline
&AVG$_{ambig}$& 0.63&	0.85&	0.68&	0.66&	0.82&	0.67 &  0.61&	0.89&	0.69&	0.63& 0.86&	\textbf{0.68}\\
&AVG$_{all}$& 0.86	&0.84	&0.81	&0.87&	0.81&	0.79& 0.85	&0.89&	0.83&	0.86&	0.86&	0.82\\
\hline
\end{tabular}
\caption{\small The results of the first method for the words given in Table \ref{tab:words}.
 The first six columns show the results of the  experiment when a vector is assigned to the closest cluster   with respect to the average similarity to all cluster members. The next six columns  report results when we choose the cluster with a vector which is closest to the one being analysed.  The columns containing the precision, recall and F1 measure, calculated using the weights {\it 1/n} and {\it 1/nc}, where {\it n} and {\it nc} denote the number of the word occurrences and the number of obtained groups, respectively.  If something is Underlined, it indicates that one of the weighting method gave higher results. The results in bold are those which are significantly (and visibly) better than the respective results from the complementary experiment within this method. Boxes show the best results from both methods (from Tables \ref{tab:1} and \ref{tab:2}). The last two rows contain results obtained for all annotated words ({\sc avg}$_{all}$) and for ambiguous words only ({\sc avg}$_{ambig}$).}
 
\label{tab:1}
\end{small}
\end{table*}

\section{Word Equivalents Prediction}
 The second method was inspired by  the idea described in \cite{amrami-goldberg-2018-word}. In this approach, we do not cluster word representations but vectors 
containing information about its possible replacements or coordinate elements within  the sentence. 
 These occurrence-based vectors, 
 called  \emph{representative vectors}, represent subsets of words predicted as being the most probable replacements for the disambiguated word
 (apart from this word itself, of course) by a contextual neural language model.
 In the original work,  for similar word prediction, an ELMo model for English was used. Since previous experiments \cite{ourgwc:2020} have shown that the Polish ELMo helps little if at all, to deal with the WSD task, we used two different  transformer-based BERT models.
 
The representative vectors are built in several steps. 
  Each coordinate of such a vector for the word $w$ corresponds to a potential substitute/representative which could be used instead (or beside it) in the sentence. 
Vectors are created separately for each part of speech to which a word may belong (i.e. we assume that we disambiguate text which is already tagged), 
based on the context of $w$ within the text. Predictions of the substitutes are performed with BERT masked model for a given context. 
In the original solution, right and left contexts were analyzed separately. Replacing the ELMo model with a transformer-based BERT model raises the natural question of whether this decision is still appropriate.
The BERT masked model allows us to predict words by considering both contexts simultaneously, which should more accurately reflect the meaning of a given word occurrence, which in turn should produce better grouping results. To check whether this conclusion is sound, we performed two sets of experiments. 
As a context we can treat  the entire sentence (\emph {both-sides context} approach) or we can use left and right contexts separately (\emph{one-side context} approach. In our work, we used both methods.
  In the \emph{one-side context} approach $k$ representatives are predicted for each of the left and right contexts separately and then $l$ elements are randomly selected $r$ times from each of the created sets. In \emph {both-sides} context approach, representatives  are predicted on the basis of the entire sentence context. From $k$ representatives $2l$ elements are drawn $r$ times.
   All predictions for a given word are added to the representative space,  common for all its occurrences.   The final size of this set defines the length of the representative vectors. 
 Each representative vector, therefore, contains at most $2l$ non zero values (some predictions can reappear for left and right contexts).
  {In this work, parameters were set to \textit{k=20},  and  \textit{r=20}.    The value of the  $l$ was set to 4 for the \emph{one-side} context approach  and \textit{l=6} for the \emph{both-sides} context approach}.
  We used two Polish BERT models --- Polbert: {\small\url{https://huggingface.co/dkleczek/bert-base-polish-uncased-v1}} and HerBERT: {\small\url{https://huggingface.co/allegro/herbert-large-cased}}, \cite{mroczkowski-etal-2021-herbert} with the Python transformers package \cite{Wolf2019HuggingFacesTS}. 
 
  For the \emph{one-side} approach, two variants of the experiment were conducted, differing in the sentence preprocessing (let $w_{i}$ be a word being disambiguated):
 
 \begin{enumerate}
 \item  \emph{AND-pattern}: word $w_i$ and the conjunction `i' (`and') were appended so 
     the schema of input data were as follows: $(w_1, w_2\ldots w_{i}, \textrm{i}, \textrm{\small\sc  <mask>})$ and $( \textrm{\small\sc  <mask>},\textrm{i}, w_i, w_{i+1}\ldots w_{n})$.
     
\item \emph{Substitution-pattern}: word $w_i$  was excluded from the context and only the following or preceding words were used for the right and left context, respectively. In this case, the schema of the input data were:  $(w_1, w_2\ldots w_{i-1}, \textrm{\small\sc  <mask>})$ and $( \textrm{\small\sc  <mask>}, w_{i+1}\ldots w_{n})$ 
  
 \end{enumerate}
  
In the second variant, for words at the beginning or end of the sentence, only the right or left non-empty context was used to find $2l$ representatives. For the \emph{both-sides} approach we used only the \emph{substitution-pattern}.

For each of the $m$ occurrences of $w$, $r$ representative vectors were clustered using the Chinese whispers algorithm.  In the constructed graph, the cosine similarity between nodes was interpreted as an arc weight. Additionally, before clustering, the matrix $mr$ $\times$ $2l$ was transformed by the TF-IDF method (each representative vector is treated as a separate document). This step was performed to reduce the weight of the words appearing in a considerable number of vectors.

We will illustrate the process on two sentences. In this example, we are using the word \emph{zamek} `castle/lock', and first variant of the method.
The sentences are presented below:

\smallskip
\begin{small}
\begin{itemize}
  \item [a.]{\small  \it Święty Bernard objeżdżał zaś miasta, zamki i wsie we Francji oraz Niemczech wygłaszając znakomite i porywające serca kazania ...} 

{\small `Saint Bernard traveled around towns, castles and villages in France and Germany, delivering excellent and thrilling sermons... } 

  \item [b.]{\small \it Wszelkie najbardziej ulepszone zamki  niewiele tu pomagają, technika złodziejska bowiem rozwija się stale, drwiąc sobie ze wszelkich zamknięć.}

{\small `All the most improved locks are of little use here, as the thieving technique is constantly evolving, mocking all locks.'}
\end{itemize}
\end{small} 

For each of the  occurrences of \emph{zamek}, from 20 predictions 4 were chosen, consecutively for left and right context, which gave us 2643 different forms. This is the final length of the representative vectors for this word.  Their elements are set initially to 0, 1, or 2 and are  further transformed by the TF/IDF coefficient.
The exemplary drawn predictions for the sentences cited above are the following:

\smallskip
\begin{small}
\begin{itemize}
  \item[a.] {\small  \emph{miasta, domy, bramy, ulice, miast, wioski, Niemcy}}
\\ {\small `towns, houses, gates, streets, towns, cottages, Germany'}

\item[b.]{\small  \emph{materiały, systemy, alarmy, technologie, lustra, złodzieje, gry, drzwi}}

 {\small `materials, systems, alarms, technologies, mirrors, thieves, games, doors'}

\end{itemize}
\end{small}


The main parts of the algorithm (predicting  representatives for all word occurrences, and clustering of representatives vectors), are computationally quite demanding.  The processing cost increases with the size of the corpus as we have to use a language model for predicting representatives for each word occurrence separately, and then we cluster as many graphs as there are word types to be disambiguated (with an algorithm of a linear cost). To limit processing time, we did not process the entire NKJP corpus,  but only 1,000 occurrences for all words to be disambiguated. 

In the final processing stage, a particular word occurrence was mapped to a cluster to which the highest number of its representative vectors  were assigned (when two groups were equally numerous, a random choice was made).  

The evaluation was done on the same text manually annotated with plWordNet senses as before. The annotated part is contained within the large data set and annotated examples are clustered together with other occurrences (at no stage of the process is information from the sense labels used). The correctness of the sense partition was checked on the basis of the annotated part.  The results of 
the \emph{one-side} context approach with Polbert language model 
for words listed in Table \ref{tab:words} are given in Table \ref{tab:2}. We used the same measures counted for every cluster member as in Table 2, limited to the annotated elements within the clusters.
We can observe there that generally \emph{substitution-pattern} outperforms \emph{AND-pattern}. Thus, in Table \ref{tab:herbert} we show the results of the experiments with \emph{both-side} approach with Polbert and Herbert language models only for this pattern.

\begin{table*}[ht!]
  \setlength{\tabcolsep}{6pt}
 \begin{small}
 \begin{tabular}{rl|rrr|rrr||rrr|rrrll}
 \hline
 && \multicolumn{12}{c}{\bf Method II: word equivalents prediction,  \emph{one-side} context approach, Polbert model}\\
 && \multicolumn{6}{c||}{\bf `AND-pattern' }& \multicolumn{6}{c}{\bf substitution-pattern}\\
  & &     P$_n$  & R$_n$  &  F1$_n$  &  P$_{nc}$  &  R$_{nc}$  &  F1$_{nc}$  &P$_n$  & R$_n$  &  F1$_n$  &  P$_{nc}$  &  R$_{nc}$  &  F1$_{nc}$      \\
  \hline
1&badanie     &  0.32  &  0.61  &  0.42  &  0.33  &  0.61  &  {0.43}  &  0.63  &  0.54  &  \fbox{\bf{0.58}}  &  0.88  &  0.39  &  0.54    \\
2&biuro    &  0.57  &  0.40  &  \bf{0.47}  &  0.62  &  0.36  &  0.46 &  0.83  &  0.26  &  0.40  &  0.91  &  0.24  &  0.38\\
3&blok     &  0.80  &  0.60  &  \fbox{\bf{0.69}}  &  0.88  &  0.56  &  0.68 &  0.8  &  0.6  & \bf{ 0.69}  &  0.88  &  0.56  &  0.68 \\
4&głos     &  0.58  &  0.46  & 0.51  &  0.75  &  0.44  &  \underline{\bf{ 0.55}}   &  0.58  &  0.41  &  \underline{0.48}  &  0.85  &  0.24  &  0.37  \\
5&historia   &  0.50  &  0.52  &  \bf{0.51}  &  0.48  &  0.48  &  0.48  &  0.82  &  0.34  &  \underline{0.48}  &  0.93  &  0.28  &  0.43  \\
6&interes    &  0.60  &  0.56  &  \bf{0.58}  &  0.76  &  0.45  &  \bf{0.57}   &  0.86  &  0.37  &  0.52  &  0.95  &  0.31  &  0.47 \\
7&język     &  0.49  &  0.67  &  \bf{\underline{0.57}}  &  0.51  &  0.58  &  {0.54} &  0.67  &  0.44  &  0.53  &  0.9  &  0.26  &  0.40    \\
8&kierunek    &  0.77  &  0.61  &  0.68  &  0.75  &  0.61  &  0.67  &  0.86  &  {0.75}  &  \fbox{\bf{0.80}}  &  0.94  &  0.67  &  \bf{0.78}    \\
9&klasa    &  0.31  &  1  &  0.47  &  0.31  &  1  &  0.47   &  1  &  0.57  &  \fbox{\bf{0.73}}  &  1  &  0.57  &  \fbox{\bf{0.73}}\\
10&kolej     &  0.38  &  1  &  0.55  &  0.38  &  1  &  0.55  &  1  &  0.75  &  \fbox{\bf{0.86}}  &  1  &  0.75  &  \fbox{\bf{0.86}}   \\
11&koło     &  0.75  &  0.51  &  0.61  &  0.83  &  0.47  &  0.60   &  0.88  &  0.56  &  \fbox{\bf\underline{0.68}}  &  0.95  &  0.44  &  0.60 \\
12&komórka    &  0.50  &  0.52  &  0.51  &  0.63  &  0.50  & \underline{0.56}  &  0.86  &  0.43  &  {0.57}  &  0.92  &  0.43  &  \fbox{\bf{0.59}} \\
13&linia     &  0.60  &  0.80  &  0.69  &  0.67  &  0.83  & \underline{0.74}  &  0.73  &  1  &  0.84  &  0.85  &  1  &  \fbox{\underline{\bf{0.92}}} \\
14&zamek     &  0.78  &  0.75  &  0.76  &  0.78  &  0.75  &  0.76  &  1  &  0.83  &  \fbox{\underline{\bf{0.91}}}  &  1  &  0.67  &  0.80  \\
\hline
& {\sc avg}$_{noun}$ & 0.57&	0.64&	0.57&	0.62&	0.62&	0.58&	0.82&	0.56&	\fbox{\underline{\bf{0.65}}}&	0.93&	0.49&	0.61
\\
\hline
15&biały     &  0.90  &  0.22  &  \bf\underline{0.35}  &  0.93  &  0.18  &  0.30  &   1  &  0.14  &  0.25  &  1  &  0.14  &  0.25  \\
16&ciężki     &  0.50  &  0.44  &  0.47  &  0.66  &  0.38  &  0.48  & 0.93  &  0.47  &  \bf{0.62}  &  0.96  &  0.47  & \fbox{\bf{0.63}} \\
17 &łagodny & 0.6 & 0.8 & 0.69 & 0.78 & 0.78 & \underline{0.78}& 1.0 & 0.8 & \bf{0.89} & 1.0 & 0.8 & \bf{0.89}  \\
18&niski  & 0.73 & 0.45 & 0.56 & 0.81 & 0.42 & 0.56 & 0.91 & 0.45 &\bf{0.61} & 0.95 & 0.45 & \bf{0.61}  \\
19&prawdziwy &0.47 & 0.55 & 0.51 & 0.57 & 0.45 & 0.5 & 0.77 & 0.54 & \fbox{\underline{\bf{0.76}}} & 0.54 & 0.63 & 0.63 \\
20&szybki & 0.62 & 0.34 & 0.44 & 0.79 & 0.31 & \bf{0.45} &  0.81 & 0.25 & 0.38 & 0.89 & 0.22 & 0.36 \\
21&zły & 0.6 & 0.43 & 0.5 & 0.69 & 0.46 & \underline{\bf{0.55}} & 0.88 & 0.35 & 0.5 & 0.93 & 0.35 & 0.51 \\
\hline
& {\sc avg}$_{adj}$& 0.63&	0.46&	0.50&	0.75&	0.43&	{0.52}&	0.85	&0.46 &	0.54 &	0.93 &	0.42 &\bf{0.55}\\

\hline
22&działać &0.64 & 0.43 & 0.51 & 0.8 & 0.35 & 0.49  &  0.73 & 0.47 & \fbox{\underline{\bf{0.57}}} & 0.95 & 0.3 & 0.46 \\
23&należeć & 0.43 & 1.0 & 0.6 & 0.43 & 1.0 & 0.6 & 0.46 & 0.93 & 0.62 & 0.72 & 0.55 & \fbox{\bf{0.63}} \\
24&przyjąć & 0.58 & 0.46 & 0.51 & 0.78 & 0.5 & \underline{0.61} & 0.67 & 0.46 & 0.55 & 0.89 & 0.48 & \underline{\bf{0.62}} \\
25&uważać & 0.79 & 0.91 & \fbox{\underline{\bf{0.84}}} & 0.93 & 0.43 & 0.59 & 0.79 & 0.8 & \underline{0.79} & 0.94 & 0.32 & 0.48 \\
26&zakładać &0.8 & 0.6 & 0.69 & 0.88 & 0.62 & \fbox{\bf{0.73}} &0.5 & 0.8 & 0.62 & 0.69 & 0.69 & \underline{0.69} \\
27&zamykać &0.77 & 0.83 & \fbox{\bf\underline{0.80}} & 0.69 & 0.81 & 0.74  & 0.75 & 0.57 & 0.65 & 0.88 & 0.53 & 0.66  \\
28&zobaczyć & 0.38 & 0.83 & 0.52 & 0.42 & 0.78 & 0.54  & 0.46 & 0.71 & 0.56 & 0.76 & 0.55 & \fbox{\bf\underline{0.64}} \\
29&związać & 1.0 & 0.29 & 0.44 & 1.0 & 0.29 & 0.44 & 0.66 & 0.6 & \underline{\bf{0.62}} & 0.84 & 0.43 & 0.57   \\
30&żyć & 0.56 & 0.6 & \underline{0.58} & 0.73 & 0.42 & 0.53 & 1.0 & 0.71 & 0.76 & 1.0 & 0.86 & \fbox{\bf\underline{0.86}} \\ 
\hline
& {\sc avg$_{verb}$}& 0.66&	0.66&	0.61&	0.74&	0.58&	0.59&	0.70&	0.65&	\fbox{\underline{\bf{0.65}}}&	0.85&	0.52&	0.55\\
\hline
&{AVG$_{30}$} & 0.61&	0.61&	0.57&	0.69&	0.56&	0.55&	0.79&	0.57&	\fbox{\underline{\bf{0.62}}}&	0.89&	0.50&	0.54\\
\hline
\end{tabular}
\caption{ The results of the second method. The first six columns show the results of the  experiment when the 'i' (`and') conjunction and an additional unknown word was added. The next six columns  report results when  a replacement of a given word was searched for on the basis of actual left and right contexts only. {\it n} and {\it nc} meanings and emphasizing rules are the same as in Table \ref{tab:1}.
Boxes show the best results from both methods (from Tables \ref{tab:1} and \ref{tab:2}).}
\label{tab:2}
\end{small}
\end{table*}

\begin{table*}[ht!]
  \setlength{\tabcolsep}{6pt}
 \begin{small}
 \centering
 \begin{tabular}{rl|rrr||rrr}
\hline
 && \multicolumn{6}{c}{\bf Method II: word equivalents prediction}\\ &&\multicolumn{6}{c}{\bf{substitution-pattern}, {\bf both-sides approach} }\\
  &&\multicolumn{3}{c||}{\bf Polbert } & \multicolumn{3}{c}{\bf HerBERT }\\

  & &       P$_{nc}$  &  R$_{nc}$  &  F1$_{nc}$ &   P$_{nc}$  &  R$_{nc}$  &  F1$_{nc}$      \\
  \hline
1	&	badanie	&	0.31	&	0.57	&	0.40	&	0.79	&	0.44	&	\textbf{0.57}	\\
2	&	biuro	&	0.88	&	0.42	&	\textbf{0.57}	&	0.84	&	0.3	&	0.44	\\
3	&	blok	&	0.67	&	0.58	&	0.62	&	0.67	&	0.58	&	0.62	\\
4	&	głos	&	0.9	&	0.25	&	0.39	&	0.83	&	0.35	&	\textbf{0.49}	\\
5	&	historia	&	0.88	&	0.29	&	0.43	&	0.83	&	0.36	&	\textbf{0.51}	\\
6	&	interes	&	0.9	&	0.39	&	0.55	&	0.94	&	0.41	&	\textbf{0.57}	\\
7	&	język	&	0.86	&	0.38	&	\textbf{0.52}	&	0.88	&	0.32	&	0.47	\\
8	&	kierunek	&	0.93	&	0.68	&	\textbf{0.79}	&	0.73	&	0.59	&	0.65	\\
9	&	klasa	&	0.71	&	0.75	&	\textbf{0.73}	&	0.67	&	0.71	&	0.69	\\
10	&	kolej	&	1.0	&	1.0	&	1.0	&	1.0	&	1.0	&	1.0	\\
11	&	koło	&	0.88	&	0.42	&	\textbf{0.56}	&	0.93	&	0.30	&	0.46	\\
12	&	komórka	&	1.0	&	0.6	&	\textbf{0.75}	&	1.0	&	0.43	&	0.60	\\
13	&	linia	&	0.88	&	0.75	&	0.81	&	1.0	&	0.80	&	\textbf{0.89}	\\
14	&	zamek	&	1.0	&	0.5	&	0.67	&	1.0	&	0.5	&	0.67	\\

\hline
& {\sc avg}$_{noun}$	& 0.84	&	0.54	&	\textbf{0.63}	&	0.87	&	0.51	&	0.62	\\

\hline
15	&	biały	&	0.93	&	0.18	&	0.30	&	1.0	&	0.22	&	\textbf{0.37}	\\
16	&	ciężki	&	0.77	&	0.39	&	0.51	&	0.9	&	0.43	&	\textbf{0.58}	\\
17	&	łagodny	&	0.88	&	0.81	&	0.84	&	0.88	&	0.81	&	0.84	\\
18	&	niski	&	0.81	&	0.47	&	\textbf{0.59}	&	0.81	&	0.44	&	0.57	\\
19	&	prawdziwy	&	0.70	&	0.42	&	0.53	&	0.77	&	0.44	&	\textbf{0.56}	\\
20	&	szybki	&	0.92	&	0.22	&	0.36	&	1.0	&	0.27	&	\textbf{0.42}	\\
21	&	zły	&	0.72	&	0.42	&	\textbf{0.53}	&	0.72	&	0.31	&	0.43	\\

\hline
& {\sc avg}$_{adj}$	&	0.82	&	0.42	&	0.52	&	0.87	&	0.42	&	\textbf{0.54}	\\

\hline
22	&	działać	&	0.94	&	0.38	&	\textbf{0.54}	&	0.95	&	0.3	&	0.46	\\
23	&	należeć	&	0.7	&	0.68	&	\textbf{0.69}	&	0.81	&	0.42	&	0.55	\\
24	&	przyjąć	&	0.82	&	0.46	&	\textbf{0.59}	&	0.84	&	0.45	&	0.58	\\
25	&	uważać	&	1.0	&	0.4	&	\textbf{0.57}	&	0.95	&	0.26	&	0.41	\\
26	&	zakładać	&	0.83	&	0.75	&	\textbf{0.79}	&	1.0	&	0.6	&	0.75	\\
27	&	zamykać	&	0.8	&	0.58	&	0.67	&	0.82	&	0.69	&	\textbf{0.75}	\\
28	&	zobaczyć	&	0.76	&	0.55	&	0.64	&	0.84	&	0.65	&	\textbf{0.73}	\\
29	&	związać	&	0.51	&	1.0	&	0.68	&	0.51	&	1.0	&	0.68	\\
30	&	żyć	&	0.76	&	0.63	&	0.69	&	0.9	&	0.76	&	\textbf{0.82}	\\

\hline
& {\sc avg$_{verb}$}	&	0.79	&	0.60	&	\textbf{0.65}	&	0.85	&	0.57	&	0.64	\\

\hline
&{AVG$_{30}$} &	0.82	&	0.53	&	\textbf{0.61}	&	0.86	&0.50	&	0.60	\\

\hline
\end{tabular}
\caption{The results of the second method using \emph{both-sides} context approach and \emph{substitution pattern} with two Bert models.   {\it nc} meaning is the same  as in Table \ref{tab:1}.} \label{tab:herbert}

\end{small}
\end{table*}

\begin{table*}[ht!]
\begin{small}
\begin{tabular}{l|l|l||lll|lll| lll|lll}
\hline
            & baseline & baseline    & \multicolumn{3}{c|}{method I. average}   & \multicolumn{3}{c|}{II. \emph{one-side} } & \multicolumn{3}{c|}{II. \emph{both-sides} } & \multicolumn{3}{c}{II. \emph{both-sides} } \\
                        &     Wordnet & max frq  & \multicolumn{3}{c|}{similarity} & \multicolumn{3}{c|}{Polbert} & \multicolumn{3}{c|}{ Polbert} & \multicolumn{3}{c}{HerBERT} \\
            \hline
            
            & F1 & F1    & P$_{nc}$  & R$_{nc}$ & F1$_{nc}$   & P$_{nc}$  & R$_{nc}$ & F1$_{nc}$  & P$_{nc}$  & R$_{nc}$ & F1$_{nc}$    & P$_{nc}$ & R$_{nc}$ & F1$_{nc}$   \\
            \hline
nouns       &0.39 &\textbf{0.65} &  0.59 & 0.76 & 0.55& 0.93       & 0.49      & {0.61} & 0.84       & 0.54      & \textbf{0.63} & 0.87      & 0.51      & {0.62} \\
adjectives  &0.38  & 0.50 & 0.58 & 0.69 & \textbf{0.62}         & 0.93       & 0.42      & {0.55} & 0.82       & 0.42   & {0.60}     & 0.87      & 0.42      & {0.54} \\
verbs   &   0.44 &  \textbf{0.73}& 0.53 & 0.79 & 0.59 & 0.56          & 0.85       & 0.52      & {0.55} & 0.79 & 0.60 & 0.85 & 0.57 & \textbf{0.64} \\
\hline
\end{tabular}
\caption{The overall results of the proposed methods and the chosen baselines for three grammatical categories.}\label{tab:overall}
\end{small}
\end{table*}

\section{Results Analysis}

The results depicted in Tables \ref{tab:1} and \ref{tab:2} do not show a clear predominance of either of the tested methods in all the cases, but the second method gives better results more often and is also better, on average, for verbs and nouns, while the first one is better for adjectives. For 30 tested words, the best solution in 19 cases was obtained by one of the variants of the second method. In 15 of these cases, the simple substitution approach was better while using coordinated phrases gave better results in 4 cases only. This finding is in contradiction with that reported in \cite{amrami-goldberg-2018-word} for English. 
The first method typically gives fewer groups (using the same clustering algorithm) and, therefore has generally better recall than precision, while the second has better precision than recall.

As an example of a good partition we can present a noun {\it linia} ({\it line}). Its 5 occurrences were partitioned into three classes. Two classes were correctly recognized, one occurrence was grouped together with two others annotated with different senses (F1=0.95). In this wrongly assigned example, {\it they are in one line}, the word \emph{line} was interpreted in the same way as in {\it penalty area line} and {\it forest line}, which is not distinctly wrong.  Some very good results were obtained in cases when only one occurrence should be classified distinctly, but the program clustered all of them together. This is the case for the adjective {\it white} in {\it white right (wing)} and the first method. The second method, which usually suggests more classes, gave much worse results here. The negative example is a word {\it examination} for which its 13 occurrences were not consistently grouped. The results for {\it direction}, which has the same number of occurrences and one sense less, are much better.

{Table \ref{tab:herbert} presents the results of the \emph{both-sides} approach, i.e. experiment in which left and right context were taken into account simultaneously while predicting word substitutes with two language models.   
Not exactly as expected, the new results for the Polbert model are only better for 11 of the 30 words checked. The average results for nouns and adjectives are worse, only for verbs are better. On average, the results from both approaches turned out to be very similar. 
The results for the HerBERT model are  better for some words worse and for the others. 
} 

Table \ref{tab:overall} summarises the overall results for all the proposed approaches and shows their comparison to the two baselines. The first baseline assumes that each occurrence of a word is assigned its first sense in plWordNet. In the Polish wordnet, this is not necessarily the most common sense of a word, although it usually is. The second  one assigns the sense which is most frequent in our test data. The proposed solutions outperform the first baseline while the second one is outperformed only for adjectives. On average, looking to both left and right contexts at once helped but for nouns the improvement is negligible. Of the two language models, Polbert turned out to be slightly more effective for nouns and adjectives, while HerBERT was better for verbs. The best overall solution for adjectives was obtained by the method I.

\section{Conclusions}
Despite   rapid development of new NLP techniques, and the emergence of  contextual language models, word sense discrimination and disambiguation are still tasks which do not have a satisfactory solution. This is especially true for under-resourced languages. 

One of the problems with the evaluation of WSD methods is the fact that many words in texts are either unambiguous or have one predominant sense. The proper evaluation of the effectiveness of methods, therefore, depends on the selection of rare examples for testing.  In the paper, we presented an evaluation of two different approaches using two different types of language models  on the selected set of ambiguous words. Although the  second method seems to be a little more precise, the  results obtained in our experiments need further investigation and larger testing data sets.

\section*{Acknowledgements}
This work was supported by the Polish National Science Centre project 2014/15/B/ST6/05186 \emph{Compositional distributional semantic models for identification, discrimination and disambiguation of senses in Polish texts}. 

\bibliographystyle{acl_natbib}
\bibliography{wsd20.bib}

\begin{thebibliography}{20}
\expandafter\ifx\csname natexlab\endcsname\relax\def\natexlab#1{#1}\fi

\bibitem[{Agirre and Edmonds(2006)}]{agi:edm:2006}
Eneco Agirre and Philip Edmonds, editors. 2006.
\newblock \emph{Word Sense Disambiguation: Algorithms and Applications}.
\newblock Springer.

\bibitem[{Agirre et~al.(2006)Agirre, Lopez~de Lacalle, Martinez, and
  Soroa}]{eneco:06}
Eneko Agirre, Oier Lopez~de Lacalle, David Martinez, and Aitor Soroa. 2006.
\newblock \href {https://doi.org/10.3115/1610075.1610157} {Two graph-based
  algorithms for state-of-the-art wsd}.
\newblock In \emph{Proceedings of the Conference on Empirical Methods in
  Natural Language Processing (EMNLP)}.

\bibitem[{Agirre et~al.(2014)Agirre, L\'{o}pez~de Lacalle, and
  Soroa}]{agirre:14}
Eneko Agirre, Oier L\'{o}pez~de Lacalle, and Aitor Soroa. 2014.
\newblock \href {https://doi.org/10.1162/COLI_a_00164} {Random walks for
  knowledge-based word sense disambiguation}.
\newblock \emph{Comput. Linguist.}, 40(1):57–84.

\bibitem[{Amigó et~al.(2009)Amigó, Gonzalo, Artiles, and Verdejo}]{bcubed:09}
Enrique. Amigó, Julio Gonzalo, Javier Artiles, and Felisa Verdejo. 2009.
\newblock \href {https://doi.org/10.1007/s10791-008-9066-8} {A comparison of
  extrinsic clustering evaluation metrics based on formal constraints}.
\newblock \emph{Information Retrieval}, 12:461–486.

\bibitem[{Amrami and Goldberg(2018)}]{amrami-goldberg-2018-word}
Asaf Amrami and Yoav Goldberg. 2018.
\newblock \href {https://doi.org/10.18653/v1/D18-1523} {Word sense induction
  with neural bi{LM} and symmetric patterns}.
\newblock In \emph{Proceedings of the 2018 Conference on Empirical Methods in
  Natural Language Processing}, pages 4860--4867, Brussels, Belgium.
  Association for Computational Linguistics.

\bibitem[{Hajnicz(2014)}]{hajn:14b}
Elżbieta Hajnicz. 2014.
\newblock Lexico-semantic annotation of {\emph{{s}kładnica}} treebank by means
  of {\textsc{plwn}} lexical units.
\newblock In \emph{Proceedings of the 7th International WordNet Conference
  ({GWC~2014})}, pages 23--31, {T}artu, Estonia. University of Tartu.

\bibitem[{Janz and Piasecki(2019)}]{janz:pias:2019}
Arkadiusz Janz and Maciej Piasecki. 2019.
\newblock A weakly supervised word sense disambiguation for {P}olish using rich
  lexical resources.
\newblock \emph{Poznan Studies in Contemporary Linguistics}, 55(2).

\bibitem[{Kędzia et~al.(2015)Kędzia, Piasecki, and
  Orlińska}]{kedzia2015word}
Paweł Kędzia, Maciej Piasecki, and Marlena Orlińska. 2015.
\newblock Word sense disambiguation based on large scale {P}olish {CLARIN}
  heterogeneous lexical resources.
\newblock \emph{Cognitive Studies| {\'E}tudes cognitives}, 15:269--292.

\bibitem[{Maziarz et~al.(2016)Maziarz, Piasecki, Rudnicka, Szpakowicz, and
  Kędzia}]{wordnet:19}
Marek Maziarz, Maciej Piasecki, Ewa Rudnicka, Stan Szpakowicz, and Paweł
  Kędzia. 2016.
\newblock Plword-net 3.0 – a comprehensive lexical-semantic resource.
\newblock In \emph{COLING 2016, 26th International Conference on Computational
  Linguistics, Proceedings of the Conference: Technical Papers}, page
  2259–2268. ACL.

\bibitem[{Mihalcea et~al.(2004)Mihalcea, Tarau, and Figa}]{Mihalcea:2004}
Rada Mihalcea, Paul Tarau, and Elizabeth Figa. 2004.
\newblock \href {https://doi.org/10.3115/1220355.1220517} {Pagerank on semantic
  networks, with application to word sense disambiguation}.
\newblock In \emph{Proceedings of the 20th International Conference on
  Computational Linguistics}, COLING '04, Stroudsburg, PA, USA. Association for
  Computational Linguistics.

\bibitem[{Moro et~al.(2014)Moro, Raganato, and Navigli}]{babelfly:14}
Andrea Moro, Alessandro Raganato, and Roberto Navigli. 2014.
\newblock Entity linking meets word sense disambiguation: a unified approach.
\newblock \emph{Transactions of the Association for Computational Linguistics
  (TACL)}, 2:231--244.

\bibitem[{Mroczkowski et~al.(2021)Mroczkowski, Rybak, Wr{\'o}blewska, and
  Gawlik}]{mroczkowski-etal-2021-herbert}
Robert Mroczkowski, Piotr Rybak, Alina Wr{\'o}blewska, and Ireneusz Gawlik.
  2021.
\newblock \href {https://www.aclweb.org/anthology/2021.bsnlp-1.1} {{H}er{BERT}:
  Efficiently pretrained transformer-based language model for {P}olish}.
\newblock In \emph{Proceedings of the 8th Workshop on Balto-Slavic Natural
  Language Processing}, pages 1--10, Kiyv, Ukraine. Association for
  Computational Linguistics.

\bibitem[{Mykowiecka et~al.(2017)Mykowiecka, Marciniak, and
  Rychlik}]{myk:mar:ryc:17:CS}
Agnieszka Mykowiecka, Małgorzata Marciniak, and Piotr Rychlik. 2017.
\newblock \href {https://doi.org/https://doi.org/10.11649/cs.1468} {Testing
  word embeddings for {P}olish}.
\newblock \emph{Cognitive Studies / Études Cognitives}, 17:1--19.

\bibitem[{Przepiórkowski et~al.(2012)Przepiórkowski, Bańko, Górski, and
  Lewandowska-Tomaszczyk}]{nkjp11a}
Adam Przepiórkowski, Mirosław Bańko, Rafał~L. Górski, and Barbara
  Lewandowska-Tomaszczyk, editors. 2012.
\newblock \emph{Narodowy Korpus Języka Polskiego}.
\newblock Wydawnictwo Naukowe PWN, Warsaw.

\bibitem[{Raganato et~al.(2017)Raganato, Delli~Bovi, and
  Navigli}]{raganato-etal-2017-neural}
Alessandro Raganato, Claudio Delli~Bovi, and Roberto Navigli. 2017.
\newblock \href {https://doi.org/10.18653/v1/D17-1120} {Neural sequence
  learning models for word sense disambiguation}.
\newblock In \emph{Proceedings of the 2017 Conference on Empirical Methods in
  Natural Language Processing}, pages 1156--1167, Copenhagen, Denmark.
  Association for Computational Linguistics.

\bibitem[{Rutkowski et~al.(2019)Rutkowski, Rychlik, and
  Mykowiecka}]{ourgwc:2020}
Szymon Rutkowski, Piotr Rychlik, and Agnieszka Mykowiecka. 2019.
\newblock Estimating senses with sets of lexically related words for {P}olish
  word sense disambiguation.
\newblock In \emph{Proceedings of the 10th Global WordNet Conference: July
  23-27, 2019, Wroclaw (Poland)}. Oficyna Wydawnicza Politechniki
  Wrocławskiej.

\bibitem[{Wawer and Mykowiecka(2017)}]{waw:myk:17:Sense}
Aleksander Wawer and Agnieszka Mykowiecka. 2017.
\newblock Supervised and unsupervised word sense disambiguation on word
  embedding vectors of unambigous synonyms.
\newblock In \emph{Proceedings of the 1st Workshop on Sense, Concept and Entity
  Representations and their Applications}, pages 120--125. Association for
  Computational Linguistics.

\bibitem[{Wolf et~al.(2019)Wolf, Debut, Sanh, Chaumond, Delangue, Moi, Cistac,
  Rault, Louf, Funtowicz, Davison, Shleifer, von Platen, Ma, Jernite, Plu, Xu,
  Scao, Gugger, Drame, Lhoest, and Rush}]{Wolf2019HuggingFacesTS}
Thomas Wolf, Lysandre Debut, Victor Sanh, Julien Chaumond, Clement Delangue,
  Anthony Moi, Pierric Cistac, Tim Rault, Rémi Louf, Morgan Funtowicz, Joe
  Davison, Sam Shleifer, Patrick von Platen, Clara Ma, Yacine Jernite, Julien
  Plu, Canwen Xu, Teven~Le Scao, Sylvain Gugger, Mariama Drame, Quentin Lhoest,
  and Alexander~M. Rush. 2019.
\newblock Huggingface's transformers: State-of-the-art natural language
  processing.
\newblock \emph{ArXiv}, abs/1910.03771.

\bibitem[{Yarowsky(1995)}]{yarowsky-1995-unsupervised}
David Yarowsky. 1995.
\newblock \href {https://doi.org/10.3115/981658.981684} {Unsupervised word
  sense disambiguation rivaling supervised methods}.
\newblock In \emph{33rd Annual Meeting of the Association for Computational
  Linguistics}, pages 189--196, Cambridge, Massachusetts, USA. Association for
  Computational Linguistics.

\bibitem[{Yuan et~al.(2016)Yuan, Richardson, Doherty, Evans, and
  Altendorf}]{yuan-etal-2016-semi}
Dayu Yuan, Julian Richardson, Ryan Doherty, Colin Evans, and Eric Altendorf.
  2016.
\newblock \href {https://www.aclweb.org/anthology/C16-1130} {Semi-supervised
  word sense disambiguation with neural models}.
\newblock In \emph{Proceedings of {COLING} 2016, the 26th International
  Conference on Computational Linguistics: Technical Papers}, pages 1374--1385,
  Osaka, Japan. The COLING 2016 Organizing Committee.

\end{thebibliography}
\end{document}